\RequirePackage[svgnames,table]{xcolor}

\documentclass[11pt,letterpaper,copyright]{style1}

\renewcommand{\copyrightext}{%
  \copyrightfont
  \textcopyright{} 2026 JP Morgan Chase \& Co. All rights reserved.%
}

\usepackage{graphicx}
\usepackage{float,epstopdf}
\usepackage{bbm}

\usepackage{microtype}

\usepackage[numbers,sort&compress]{natbib}

\usepackage{subcaption}
\usepackage{booktabs}

\usepackage{amsmath}

\usepackage{amssymb}
\usepackage{mathtools}
\usepackage{amsthm}
\usepackage{dsfont}
\usepackage{multicol}
\usepackage{makecell}
\usepackage{multirow} 
\usepackage{amsfonts} 
\usepackage{mathrsfs}
\usepackage[amssymb, thickqspace]{SIunits}
\usepackage{enumitem}
\usepackage{pgfplotstable}
\usepackage{lipsum}		% Can be removed after putting your text content

\usepackage{microtype}
\usepackage{graphicx}
\usepackage{booktabs} % for professional tables
\usepackage[table]{xcolor}
\usepackage{arydshln}
\usepackage[normalem]{ulem} % [normalem] prevents the package from changing the default behavior of `\emph` to underline.

\usepackage{cases}
\usepackage{wrapfig}

\usepackage{url}

\usepackage{thmtools}
\usepackage{thm-restate}

\definecolor{huskypurple}{HTML}{4B2E83}

\usepackage{titletoc}

\usepackage{listings}
\lstdefinestyle{promptstyle}{
  basicstyle=\ttfamily\footnotesize,
  breaklines=true,
  breakautoindent=false,
  breakindent=0pt,
  postbreak=\mbox{\textcolor{gray}{$\hookrightarrow$}\space},
  columns=fullflexible,
  keepspaces=true,
  frame=single,
  framesep=5pt,
  xleftmargin=6pt,
  xrightmargin=6pt,
  aboveskip=8pt,
  belowskip=8pt,
  showstringspaces=false,
}   % third-party packages (natbib, booktabs, listings, ...)
\makeatletter
\def\munderbar#1{\underline{\sbox\tw@{$#1$}\dp\tw@\z@\box\tw@}}
\makeatother

\AddToHook{cmd/appendix/before}{%
  \setcounter{axiom}{0}%
}

\newcommand{\be}{\begin{equation}}
\newcommand{\ee}{\end{equation}}
\newcommand{\bea}{\begin{equation*}\begin{aligned}}
\newcommand{\eea}{\end{aligned}\end{equation*}}

\definecolor{FMAccent}{HTML}{8F5A39}
\definecolor{FMTint}{HTML}{F8F4F1}
\definecolor{FMTable}{HTML}{FEF4DB}

\renewcommand{\titlefont}{\centering\color{FMAccent}\normalfont\bfseries\fontsize{18}{20}\selectfont}

\newtcolorbox{FMDefinition}{
    enhanced,
    breakable,
    colback=White,
    frame hidden,
    borderline west={2pt}{0pt}{FMAccent},
    sharp corners,
    boxsep=0pt,
    left=10pt, right=0pt,
    top=2pt, bottom=2pt,
    before skip=10pt,
    after skip=10pt,
}

\newtcolorbox{FMOverview}[1]{
    enhanced,
    breakable,
    title={#1},
    colback=FMTint,
    colframe=FMAccent!55!White,
    colbacktitle=FMAccent,
    coltitle=White,
    fonttitle=\normalsize\bfseries,
    titlerule=0pt,
    boxrule=0.5pt,
    arc=2mm,
    boxsep=0pt,
    left=9pt, right=9pt,
    top=7pt, bottom=7pt,
    toptitle=5pt, bottomtitle=5pt,
    before skip=10pt,
    after skip=12pt,
}

\newtcolorbox{FMBet}{
    enhanced,
    breakable,
    colback=FMTint,
    colframe=FMAccent!55!White,
    boxrule=0.5pt,
    arc=2mm,
    boxsep=0pt,
    left=7pt, right=7pt,
    top=5pt, bottom=5pt,
    before skip=8pt,
    after skip=9pt,
}
\newcommand{\FMBetLabel}[1]{\raisebox{0.12ex}{\textcolor{FMAccent}{\textbf{#1}}}}

\tcbset{
    titlebox/.style={
        colback=FMTint,
        colframe=FMAccent!8!White,
        boxrule=0.5mm,
        arc=2mm,
        auto outer arc,
        left=5mm,
        right=5mm,
        top=5mm,
        bottom=5mm,
    }
}

\newtcolorbox{simpleElegantQuote}{
    colback=FMTint,
    colframe=FMAccent,
    boxrule=0.5pt,
    arc=2mm,
    boxsep=4pt,
    left=10pt, right=10pt,
    top=8pt, bottom=8pt,
    fontupper=\itshape,
}

\title{Toward Structured Financial Foundation Models}

\runningtitle{Toward Structured Financial Foundation Models - JPMorgan AI Research}


\usepackage{fontawesome5}   % brand/utility icons, e.g. \faGithub

\author{Natraj Raman\textsuperscript{*}, Saumitra Mishra, Vamsi Potluru and Sumitra Ganesh\\
JPMorgan AI Research\\
}

\hypersetup{
    pdftitle={Toward Structured Financial Foundation Models},
    pdfauthor={Natraj Raman},
    pdfsubject={A research agenda for structured financial foundation models: persistent state, learned interoperability, and action-conditioned simulation.},
    pdfkeywords={structured data, domain models, representation interoperability, dynamics, simulation},
    pdfdisplaydoctitle=true,
    colorlinks=true,
    linkcolor=blue!50!black,
    citecolor=blue!50!black,
    urlcolor=blue!50!black
}

\begin{document}

\begin{abstract}
\vspace{-1mm}
{\centering\section*{Abstract}}
Foundation models promise reusable capabilities and step-change performance gains across tasks. In finance, large-scale pretraining of domain-specialized foundation models can advance both goals. However, prevailing architectures and objectives do not by themselves address the combined demands of financial systems: partial observability, interacting entities, asynchronous events, changing regimes, action-dependent outcomes, and rare states with disproportionate consequences. Moreover, independently developed specialists may learn incompatible representations, while multi-domain monolithic alternatives couple data access, model updates, and governance. We therefore propose a structured-data-first research agenda centered on modular, domain-specialized financial foundation models connected by a representation fabric for reuse and composition across domains and tasks. These models should maintain persistent financial state across aggregation levels and time horizons, capture its dynamics, and simulate action-conditioned futures, while remaining calibrated across regimes and for tail outcomes. They must also learn efficiently under data and access constraints and provide faithful, traceable explanations. We identify research challenges, directions, and capability-oriented evaluation criteria.
\end{abstract}

\maketitle

\begingroup
\renewcommand{\thefootnote}{\fnsymbol{footnote}}
\footnotetext[1]{Corresponding author: \texttt{natraj.raman@jpmorgan.com}}
\endgroup

% =============================================================================
\section{Introduction}
\label{sec:intro}
% =============================================================================

Can the paradigm of foundation models (FMs)~\citep{zhou2025comprehensive} extend to the structured data that underpin financial activity---transactions, customers, accounts, firms, securities, orders, positions, and portfolios? Financial institutions repeatedly address diverse prediction and decision problems involving interconnected entities and events~\citep{biswas2020ai}, yet their models and representations remain fragmented across datasets and tasks. The challenge is to make them reusable while substantially improving performance.

One promising approach is large-scale pretraining of modality-specific architectures---such as tabular~\citep{jiang2026representation}, time-series~\citep{wang2026deep}, or graph~\citep{liu2025graph} models---on financial data, followed by adaptation to downstream tasks. Another approach adapts language-model architectures to structured financial records through text serialization or tokenization~\citep{gruver2023large,kim2024financial,shi2026kronos}. These approaches can improve task performance and support reuse across tasks~\citep{zhu2025fincast,li2025mars}, but do not by themselves address the wider demands of financial systems.

Financial activity unfolds across complex networks of interacting entities, under changing regimes, and with rare, consequential outcomes~\citep{hayman2025reframing}. Computationally, this setting is asynchronous and partially observable, with outcomes shaped by agents' actions~\citep{ardon2021towards}. These properties require capabilities beyond those induced by prevailing architectures, representations, and objectives. Reuse also presents a system-level challenge: independently developed domain models may learn incompatible representations, while multi-domain monolithic models~\citep{wu2023bloomberggpt} couple data access, model updates, and governance.

To address these gaps, we advance a structured-data-first agenda built around domain-specialized financial FMs and learned interoperability among them. Each specialist FM focuses on a coherent area of financial activity, such as transactions, markets, credit, or portfolios, so it can evolve independently within domain boundaries. Records, events, entities, and relationships form its modelling substrate, supplemented where useful by unstructured information. State representations describe the domain's financial objects and processes, such as accounts and repayment behaviour in credit, or securities and order flow in markets, while a downstream task may draw on multiple specialists.

\begin{figure*}[t]
	\centering
	\includegraphics[width=\textwidth]{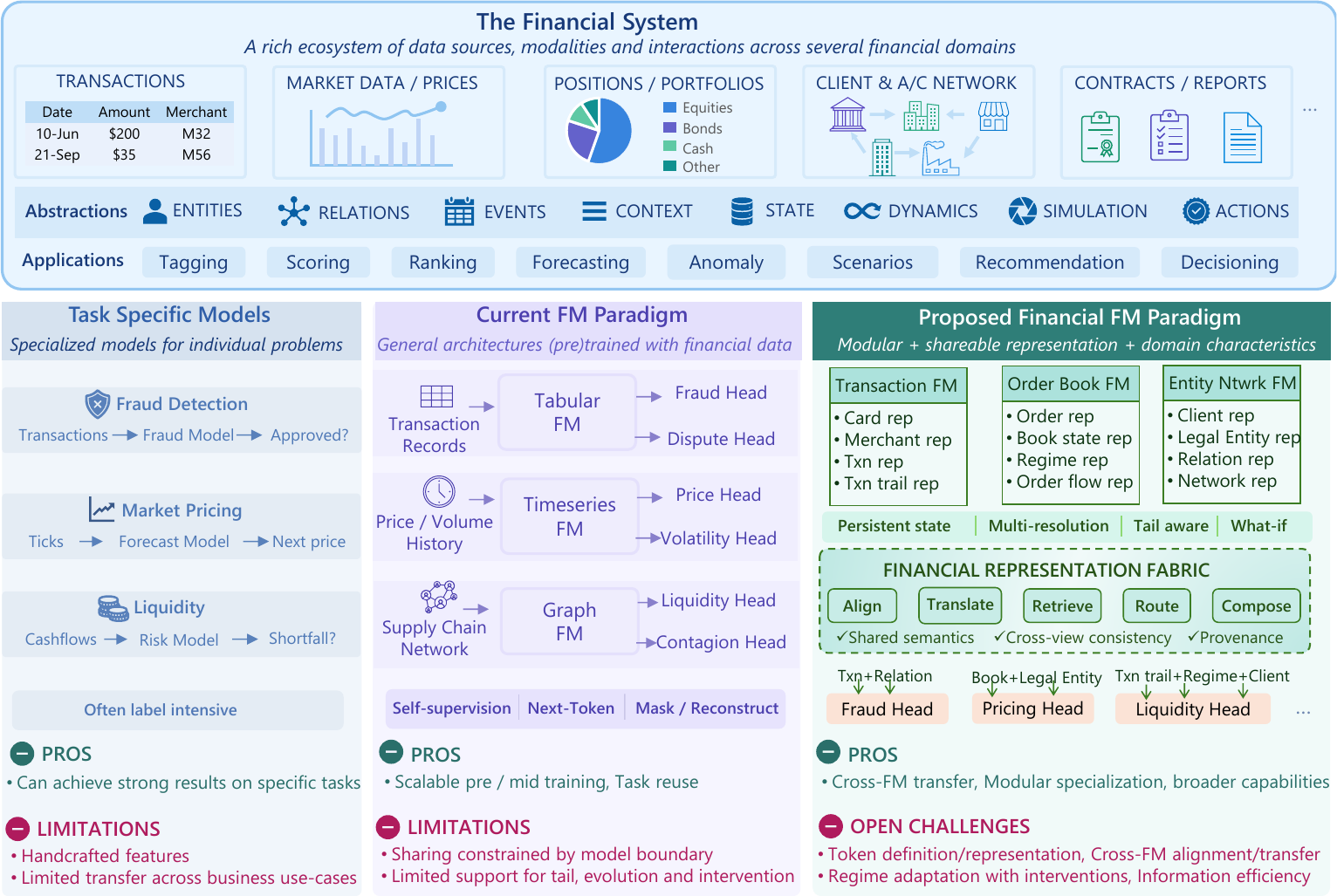}
	\caption{\textbf{From isolated models to a structured financial foundation-model paradigm}. \emph{Top}: Diverse financial domains and modalities connect through shared abstractions and support a broad range of tasks. \emph{Left}: Task-centric models learn specialized mappings for individual financial problems. \emph{Middle}: Modality-specific FMs share pretrained representations across downstream tasks. \emph{Right}: Specialist financial FMs produce interoperable, multi-resolution representations that a representation fabric can compose across model boundaries, enabling task heads to combine state from multiple FMs.}
	\label{fig:fm-overview}
\end{figure*}

\begin{FMDefinition}
\textbf{Our position.} A structured-data-first financial FM ecosystem should build on domain-specialized pretraining through specialists that construct explicit, persistent state representations for events, entities, and relationships across varying hierarchical levels and temporal resolutions, treating these representations as the primary transferable objects. Each specialist should learn regime-conditioned and action-conditioned state transitions, select evidence under data and access constraints, represent uncertainty and tails, and ground outputs in traceable evidence and state changes. A learned \emph{financial representation fabric} should enable reuse across independently trained specialists through discovery, translation, routing, and compatibility assessment, rather than requiring a single monolithic FM spanning all financial domains and use cases.
\end{FMDefinition}

The goal is not to claim a complete financial world model~\citep{chen2026definition}, but to frame a research agenda around testable hypotheses. We identify mismatches between financial-system structure and prevailing foundation-model assumptions (\S~\ref{sec:why-different}), derive required capabilities (\S~\ref{sec:desiderata}), describe the obstacles (\S~\ref{sec:challenges}), outline research directions (\S~\ref{sec:research-directions}), and propose evaluation principles (\S~\ref{sec:evaluation}).~\autoref{fig:fm-overview} summarizes the progression from isolated task models toward reusable, interoperable specialist systems.

% =============================================================================
\section{Why Finance Is Different}
\label{sec:why-different}
% =============================================================================

Conventional approaches apply an existing FM paradigm to financial data in its native modality~\citep{chen2025advancing, lu2025time, yanglet2025multimodal}, combining generic self-supervision with downstream adaptation. Matching modalities and scaling pretraining can yield substantial task-level gains. For a broader financial FM paradigm, however, an additional challenge is structural: financial observations arise from evolving systems of persistent entities, relationships, decisions, and constraints.~\autoref{tab:financial-mismatches} summarizes eight mismatches. Several also arise in other domains, but their co-occurrence---sharpened by accounting identities, strategic and adversarial participants, mandated information barriers, and direct economic consequences---distinguishes the financial setting.

\begin{table*}[t]
	\centering
	\small
	\captionsetup{justification=centering}
	\caption{Structural and institutional mismatches and the capabilities they motivate.}
	\label{tab:financial-mismatches}
	\renewcommand{\arraystretch}{1.15}
	\begin{tabular}{p{0.25\textwidth}p{0.25\textwidth}p{0.37\textwidth}}
		\toprule
		\textbf{Financial property} & \textbf{Assumption challenged} & \textbf{Required capability} \\
		\midrule
		Interacting entities & Independent observations & Relational state representation \\
		Multi-scale event streams & Regular sequences & Event-aware, multi-scale representation \\
		Regime dependence & Stable distributions & Adaptive dynamics \\
		Actions affect outcomes & Passive prediction & Action-conditioned simulation \\
		Consequential tails & Average-case objectives & Tail-aware learning \\
		Partial observability & Complete context & Uncertainty-aware state inference \\
		Vast data; sparse evidence & Abundant pooled data & Selective learning; synthetic coverage \\
		High-consequence decisions & Accuracy is sufficient & Explainable, auditable outputs \\
		\bottomrule
	\end{tabular}
\end{table*}

\subsection{Financial systems are relational, event-driven, and multi-scale}

Customers, firms, counterparties, accounts, securities, and portfolios persist through time. Events can alter several entities, and their meaning depends on identity and relationships. Independent rows discard this continuity, while homogeneous token streams---including text serializations of records---obscure which objects persist and which relationships change~\citep{bardoscia2021physics}. The useful modelling unit is the state of an interacting system that observations only partially reveal.

Observations also arrive asynchronously across milliseconds to years. A fixed sequence index obscures elapsed time and the distinction between transient activity and persistent state, while exhaustive attention over every historical event is neither computationally plausible~\citep{katharopoulos2020transformers} nor useful.

\subsection{The data-generating process changes}

Market structure, institutional constraints, and customer or participant behaviour can cause concept drift or structural breaks~\citep{gama2014survey}. Models must separate persistent relationships and constraints from changing dynamics and detect when adaptation is warranted.

Credit limits, portfolio changes, prices, and controls alter subsequent outcomes, so historical data reflect both the system and its decision rules. When deployed policies induce participants to respond or adapt, relationships learned under one policy may fail under another~\citep{lucas1976econometric}. Predicting what followed an observed state does not establish what would follow a different action~\citep{thummcausal}.

\subsection{Average performance and point predictions hide consequential failure}

Rare outcomes such as defaults, liquidity shocks, operational failures, and market dislocations contribute little to average likelihood yet can dominate financial value and risk ~\citep{taleb2020statistical}. Scaling may improve common-state modelling without correcting tail miscalibration. Moreover, point predictions can conceal uncertainty about inferred state and regime when evidence is incomplete or conflicting~\citep{saly2026deep}, materially affecting decisions about credit limits, capital allocation, or hedging.

Financial state is also partially observed: records omit intentions, off-platform activity, delayed reports, and external relationships. Models must infer latent state without becoming certain about what they cannot observe~\citep{yang2026observations}.

\subsection{Data access and model use are constrained}

Prevailing FM recipes rely either on large pooled corpora~\citep{liu2025sundial} or on synthetic distributions broad enough to transfer to real datasets~\citep{hollmann2025accurate,qu2026tabiclv2}. Neither assumption is straightforward in finance: records may be vast, but sensitivity, permissions, fragmentation, and sheer volume prevent indiscriminate pooling or use, even as evidence for task-relevant states and outcomes remains sparse. Realistic synthetic priors must also capture interacting entities, changing regimes, actions, and consequential tails.

Financial models may influence credit, fraud, trading, risk, and operational decisions subject to formal governance review. Such review requires understanding the basis and limitations of a model-assisted decision, not just its predictive accuracy~\citep{vcernevivciene2024explainable}.

\begin{FMDefinition}
\textbf{Core implication.} Together, these properties challenge the idea that the financial foundation-model problem is simply to apply a larger generic model to more financial data. They broaden the research question beyond \emph{how much data and capacity are required?} to also ask \emph{what should be represented, how should financial state evolve, and which capabilities should pretraining support?}
\end{FMDefinition}

% =============================================================================
\section{Desiderata for Structured Financial Foundation Models}
\label{sec:desiderata}
% =============================================================================

\begin{figure*}[t]
	\centering
	\includegraphics[width=\textwidth]{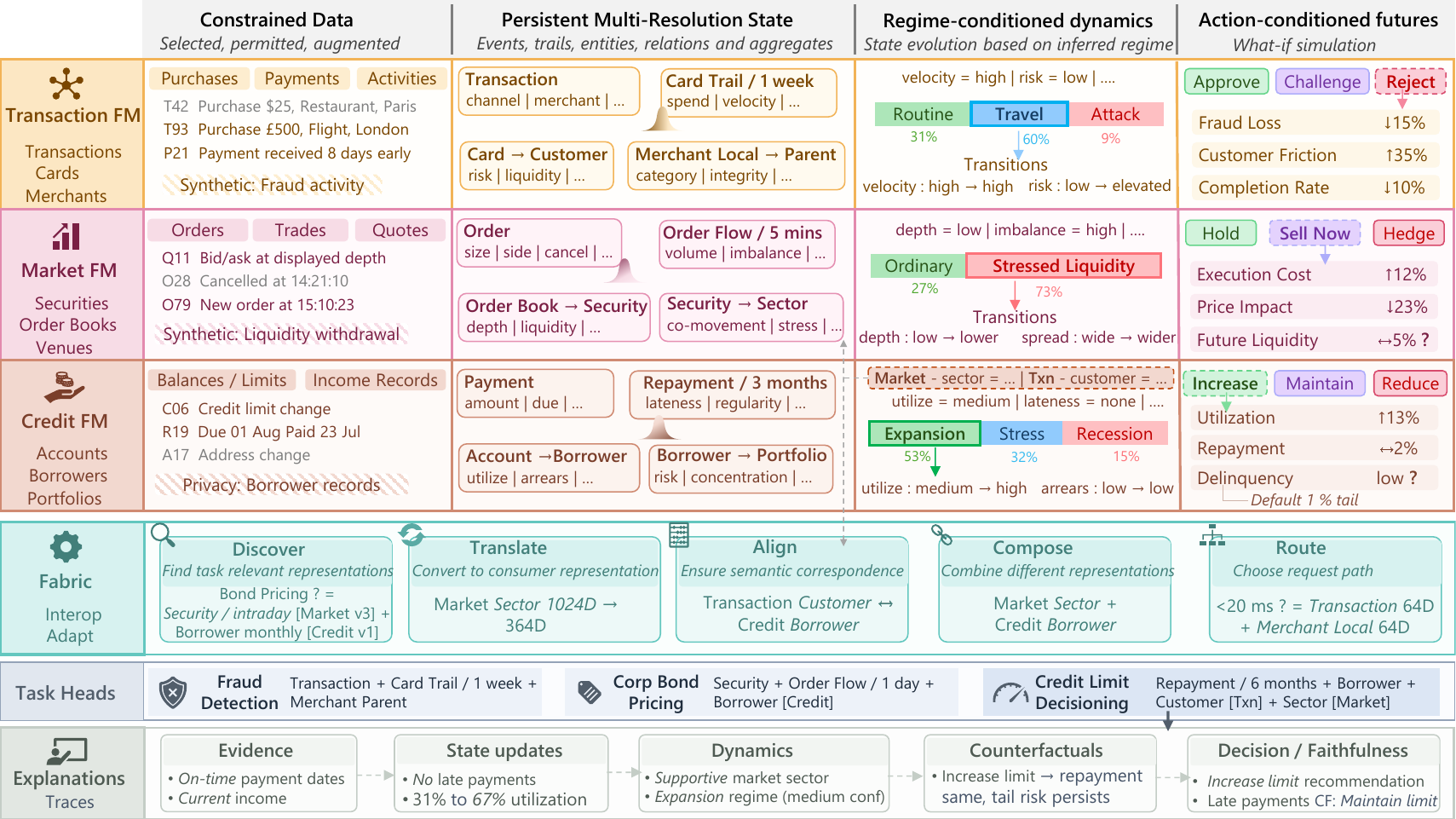}
	\caption{\textbf{Illustrative (hypothetical) capability view for structured financial foundation models.} Specialist FMs independently construct financial state representations, model their regime-conditioned dynamics, and simulate action-conditioned futures. In the cross-model example, the \emph{Credit FM} augments its native \emph{borrower} state with \emph{sector} state from the \emph{Market FM} and \emph{customer} state from the \emph{Transaction FM}, exchanged through the representation fabric. Task heads consume representations either directly from specialists (\emph{Fraud Detection}) or after adaptation and composition through the fabric (\emph{Credit Limit Decisioning}). Uncertainty is preserved and propagated throughout, while a provenance-aware trace links a decision to its evidence, state changes, and faithfulness checks.}
	\label{fig:fm-desired}
\end{figure*}

Addressing the mismatches in Section~\ref{sec:why-different} across the full paradigm requires more than scale or average downstream accuracy. We group the requirements for structured financial FMs into seven capabilities. \autoref{fig:fm-desired} illustrates how these capabilities interact within transaction, market, and credit FMs and across model boundaries.

\begin{FMOverview}{Desiderata at a glance}
\small
\begin{tabularx}{\linewidth}{@{}>{\bfseries}p{0.27\linewidth}X@{}}
Financial state & How should entities, events, and relations be represented across resolutions? \\
Dynamics & How should state evolve and adapt across regimes? \\
Actions & How should decisions alter possible futures? \\
Uncertainty and tails & What is unknown, and how should the model behave under stress? \\
Reuse and specialization & What should transfer across modular financial FMs? \\
Data and access & How should relevant data be selected, shared, or supplemented? \\
Traceability & What observations and state changes support an output? \\
\end{tabularx}
\end{FMOverview}

\paragraph{Represent meaningful financial state.}

A financial FM should represent persistent entities, their relationships, events, and partially observed latent state. It must support irregular timing, multiple temporal scales, and multiple entity and relational granularities, relating local activity to broader geographic, organizational, or portfolio-level state.

\paragraph{Model dynamics and adapt across regimes.}

State-transition modelling should distinguish stable financial structure from regime-dependent behaviour. The FM should detect changes in dynamics, localize updates to the affected behaviour, and retain knowledge that remains valid.

\paragraph{Represent actions and possible futures.}

Financial FMs should model how actions alter possible future states while preserving accounting, relational, and temporal consistency. They should distinguish conditional scenarios from causal, counterfactual, or strategic-response scenarios.

\paragraph{Treat uncertainty and tails as first-class objects.}

Uncertainty about inferred state, the active regime, and future outcomes should remain distinguishable under limited data, partial observability, conflicting signals, and distribution shift. Calibration should support reliable decisions for low-frequency, high-impact outcomes without degrading common-state performance.

\paragraph{Transfer through a representation fabric and selective specialization.}

\begin{FMDefinition}
\textbf{Financial representation fabric.} A paradigm for governed reuse of state representations across independently trained specialist FMs. Given a consumer's request for an entity, event, or relationship---specified by as-of time, temporal granularity, hierarchical level, and intended use---the fabric returns a view from permitted producers, adapted to the consumer's representation space and accompanied by provenance, uncertainty, and a compatibility estimate, or declines the request. It is not a single architecture, shared schema, representation store, or orchestration layer; its distinctive AI claim is that learned interoperability can generalize beyond bespoke fixed-pair integrations to newly paired producer--consumer combinations and model upgrades with limited adaptation.
\end{FMDefinition}

State reuse through the fabric should enable downstream adaptation with limited labelled data, parameter updates, and deployment overhead while preserving specialist capabilities.

\paragraph{Learn efficiently from constrained data.}

Pretraining should prioritize observations by state change, and inference should retrieve only relevant, permitted history and relationships, so costs scale with useful information. Training may combine synthetic, privacy-transformed, and real data to ease access constraints while preserving capability. Model checkpoints should retain source provenance and permitted uses so access and release reflect training lineage.

\paragraph{Ground outputs in evidence and state changes.}

Predictions, simulations, and recommendations should be traceable to the observations, retrieved context, representations, and state transitions that materially influenced them. Explanations should distinguish evidence from inference, attribute contributions across specialists, and be faithful, robust, auditable, concise, and intelligible.

\iffalse
\subsection{Research hypotheses}

These desiderata lead to the following hypotheses that make the position testable:

\begin{enumerate}
	\item \textbf{Multi-scale, entity- and event-aware state representations} will transfer more effectively across financial tasks than representations based on independent tables or homogeneous sequences.
	\item \textbf{A representation fabric connecting specialized financial FMs} will provide a better trade-off between reuse and task performance than either a monolithic model or fully disconnected models.
	\item \textbf{Explicit modelling of state transitions and actions} will improve generalization across regimes relative to pure next-event prediction.
	\item \textbf{Average-likelihood objectives alone} will underrepresent consequential tail states and yield unreliable uncertainty under partial observability and distribution shift.
	\item \textbf{State-informed evidence selection and trace-grounded explanations} will improve efficiency, robustness, and faithfulness over generic retrieval, unconstrained synthesis, and post-hoc explanations.
\end{enumerate}

These hypotheses admit competing architectures and objectives while defining evidence that would advance or weaken the paradigm. The next section turns to the scientific and institutional obstacles that must first be resolved.
\fi

% =============================================================================
\section{Open Challenges}
\label{sec:challenges}
% =============================================================================

The desiderata clarify the target but not how to realize it. Representation choices constrain learnable dynamics, available evidence limits claims about interventions and tails, and institutional boundaries limit reuse. The central challenges concern abstractions, evidence, and governance rather than a checklist of model components.

\paragraph{Defining the object of representation.}

Tabular and sequential FMs~\citep{sun2026survey} learn patterns from records or event histories but do not necessarily maintain persistent, addressable entity and relationship state that can be updated and reused across tasks. Graph models~\citep{fey2024position,liu2025graph} provide explicit entity--relation structure, while temporal graph models~\citep{rossi2020temporal} capture evolving interactions and can maintain entity memory. However, FM-scale pretraining and reuse raise challenges in scaling relational aggregation, accommodating heterogeneous schemas and entity types, and maintaining useful state over large, continually evolving graphs.
%Financial events differ in scale, timing, and semantics while still involving persistent entities and changing state.

\emph{Open question.} Which identities, event semantics, relationships, continuous quantities, missingness, and delayed observations should be represented, and which should remain specialist-specific? A universal vocabulary may erase task-relevant detail, while dataset-specific encodings reproduce fragmentation; the boundary between persistent state, retrieved context, and specialist computation remains open.

\begin{FMOverview}{Research gaps at a glance}
\small
\begin{tabularx}{\linewidth}{@{}>{\bfseries}p{0.27\linewidth}X@{}}
Representation & No common abstraction for heterogeneous entities, events, and state \\
Objectives & Reconstruction alone may not induce transferable financial capabilities \\
Dynamics & Persistent structure and regime-specific behaviour are entangled \\
Intervention & Observational prediction does not identify action effects \\
Uncertainty and tails & Evidence is incomplete and consequential extremes are scarce \\
Interoperability & Reuse must preserve specialization, control, and governance \\
Data and coverage & Data are vast, restricted, and sparse for plausible scenarios \\
Explanation & Outputs must remain faithful to evidence and composed state changes \\
\end{tabularx}
\end{FMOverview}

\paragraph{Learning useful objectives beyond reconstruction.}

Masked reconstruction and next-token prediction provide scalable self-supervision for structured and event data~\citep{raman2024scalable, ostroukhov2026pragma, kawawa2026tradefm}. JEPA-style objectives predict target representations rather than reconstruct observations~\citep{assran2023self}. Neither objective family necessarily yields transferable representations of persistent financial state. Direct supervision is also limited because these states are rarely observed in full.

\emph{Open question.} Which pretraining signals recover persistent state, relational structure, and multi-horizon dynamics that transfer beyond the source task? Objectives must balance fidelity and reuse: retaining too much source detail inhibits transfer, while abstracting too aggressively undermines specialist performance.

\paragraph{Separating persistent structure from changing dynamics.}

Drift-detection methods~\citep{lukats2025benchmark,amoukou2024sequential} can flag distributional changes, while continual-learning-based adaptation~\citep{guo2025comprehensive} can accommodate them. Neither establishes whether observed drift reflects noise, a temporary regime, or durable structural change~\citep{wang2026time}, particularly when financial subsystems evolve at different rates.

\emph{Open question.} What should remain invariant, what should adapt, and when does the evidence warrant an update? Regimes are rarely labelled and change points may appear only retrospectively; rapid adaptation risks forgetting stable relationships, while conservative updates preserve obsolete dynamics.

\paragraph{Distinguishing prediction from intervention.}

Causal inference and off-policy evaluation reason about interventions from observational or experimental evidence~\citep{pearl2016causal,thummcausal}, but depend on identification assumptions, overlap, or controlled variation. Financial histories show outcomes under actions shaped by policies and information that may be absent from training data, while participants may adapt to deployed policies or strategically alter observed signals.

\emph{Open question.} When do learned dynamics support forecasting, conditional simulation, policy evaluation, strategic-response modelling, or causal intervention? Action conditioning can reproduce historical associations without identifying action effects, and fixed-response simulations can miss feedback, gaming, or equilibrium change; stronger claims require corresponding identification and validation evidence.

\paragraph{Separating event likelihood, severity, and uncertainty.}

Reweighting, robust objectives, extreme-value methods, and synthetic generation target low-frequency outcomes, tail severity, or uncertainty under shift and weak support~\citep{zhang2026uncertainty,gu2025beyond,zhang2026resga}, but these are distinct problems. Crashes, defaults, fraud, liquidity shocks, and operational failures exhibit different combinations of rarity, severity, and shift: pooling them obscures distinct mechanisms, while separating them leaves little data.

\emph{Open question.} How can models disentangle rare-event likelihood, tail severity, and evidential support when extremes are scarce and mechanisms differ? Historical extremes cover only a narrow range of future stress, while synthetic cases can reproduce blind spots; models must communicate when predictions exceed evidential support.

\paragraph{Sharing knowledge without erasing boundaries.}

Multi-task learning, representation alignment, adapters, and modular routing share knowledge across models or domains~\citep{pfeiffer2023modular,yu2024unleashing}, but commonly assume joint training, a shared representation space, or mappings fixed to known models, tasks, or pairs. Operational interfaces can standardize access and metadata without establishing semantic compatibility between independently learned representations. New consumers or model versions may change representation semantics, and sharing can cause negative transfer.

\emph{Open question.} Can learned interoperability discover and translate relevant producer state for an independently trained consumer, detect conflicts among producer views, and predict incompatibility or information loss without an engineered mapping for every pair or version? The AI challenge is to preserve specialist performance across heterogeneous, evolving representation spaces within privacy, lineage, latency, cost, and model-risk constraints.

\paragraph{Selecting relevant data and expanding coverage.}

Adaptive sampling, retrieval, and long-context methods reduce learning costs~\citep{taguchi2025efficient}. Privacy-preserving methods can enable learning from restricted data~\citep{elsheikh2026privacy}, while synthetic generation can reduce direct data use and expand scenario coverage~\citep{potluru2023synthetic,buhlertowards}. In finance, however, transformations may remove entity, relational, temporal, regime, or tail information needed to infer state, while synthetic-first training may encode generator assumptions that later real-data adaptation cannot repair. Synthetic data may also leak source information or violate financial structure.

\emph{Open question.} How should observed, transformed, and synthetic evidence be selected and combined during pretraining and specialization without discarding material signals or replacing evidence with plausible synthesis? Provenance must capture timing, entitlements, generators, transformations, adaptation stages, and inherited restrictions.

\paragraph{Explaining composed financial state.}

Feature attribution, example-based explanation, concept-based methods, and provenance tracing provide partial accounts of model behaviour~\citep{bilovs2026mechanistic, balef2026one}, but they commonly explain one model or prediction at a time. A composed financial output may instead depend on retrieved evidence, latent state updates, representation translation, routing, and several independently trained specialists.

\emph{Open question.} How can an explanation identify the permitted evidence, inferred state changes, specialist contributions, and fabric operations that produced a composed output? Plausible narratives are insufficient: explanations must faithfully disclose the conditions and limitations affecting interpretation.

\begin{FMDefinition}
\textbf{Core implication.} Existing methods provide starting points, but the required capabilities do not follow automatically from their use. These gaps shift the research question from \emph{which methods address each requirement?} to \emph{what must be learned at scale so that these capabilities transfer beyond their original settings?}
\end{FMDefinition}

\iffalse
Research questions retained for possible expansion or illustration:
\begin{itemize}
	\item Semantic compatibility: Does an entity or state have the same meaning to transaction, fraud, treasury, and other domain models?
	\item Temporal validity: When was a state valid, and when did each model observe it?
	\item State ownership: Which model may update a representation, and how should conflicting updates be reconciled?
	\item Uncertainty propagation: How should confidence or unsupported state travel across composed models?
	\item Selective disclosure: Can a model share useful state without exposing restricted source data?
	\item Replaceability: Can one specialist be upgraded or replaced without retraining every consumer?
	\item Compositional failure: How should stale state, translation loss, and propagated errors be detected?
\end{itemize}
\fi

% =============================================================================
\section{Research Directions}
\label{sec:research-directions}
% =============================================================================

\begin{figure*}[t]
	\centering
	\includegraphics[width=\textwidth]{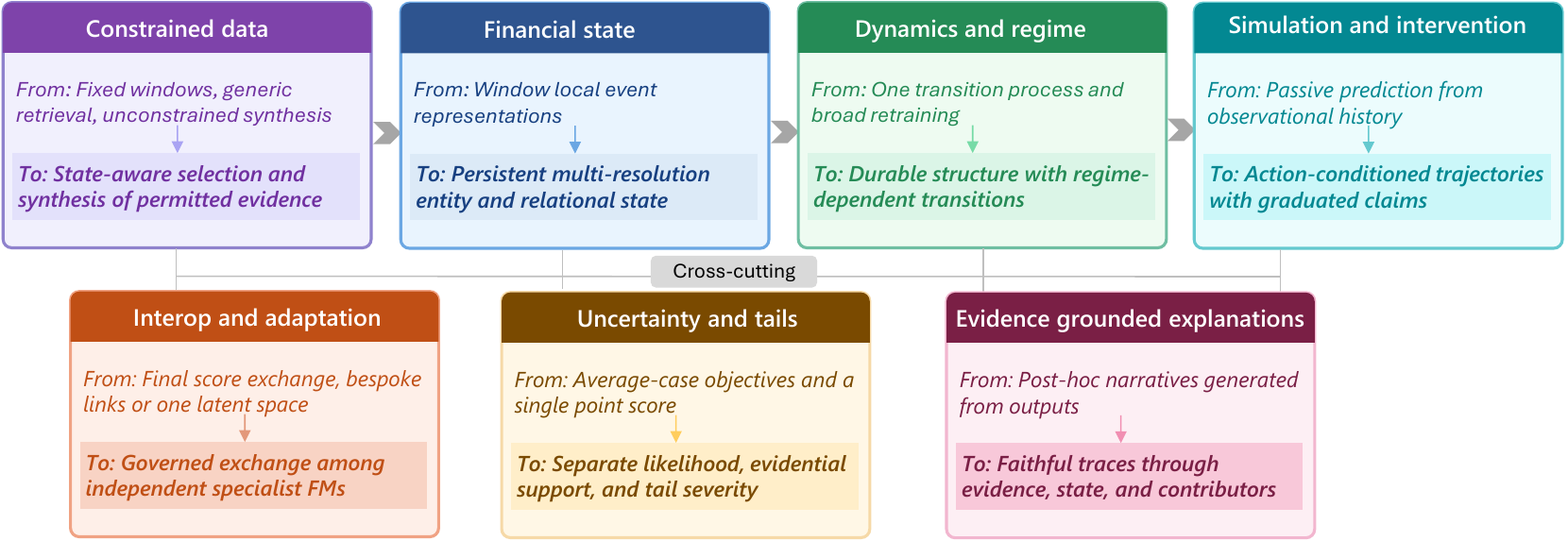}
	\caption{\textbf{Seven interconnected research bets.} Each card contrasts prevailing practice with the proposed shift. Top pane traces the operational flow. Bottom pane shows capabilities that act across it.}
	\label{fig:fm-direc}
\end{figure*}

The challenges above motivate seven research bets. Each states a claim, plausible mechanisms, trade-offs, and testable implications without prescribing an architecture. The section follows conceptual dependencies; \autoref{fig:fm-direc} shows the operational flow.

\subsection{Reusable state and interoperability}

\subsubsection{Persistent financial state as the unit of transfer}

\textbf{Research bet.} The primary transferable object should be explicit, multi-resolution entity, event, and relationship state, rather than state that remains implicit within a model's event sequence, parameters, or context window. Explicit state is addressable outside the producing FM, is updated as new observations arrive, carries uncertainty, and can be selectively consumed by new tasks and specialists. Pretraining should recover these states and their transitions rather than optimize event reconstruction alone.

\begin{FMBet}
\small
\renewcommand{\arraystretch}{1.12}
\begin{tabularx}{\linewidth}{@{}p{0.16 \linewidth}X@{}}
\FMBetLabel{Core claim} & Persistent, multi-resolution state enables reuse beyond its originating task and model. \\
\FMBetLabel{Approaches} & Typed encoding; entity-centric attention; hierarchical reconciliation; selective exports. \\
\FMBetLabel{Central tension} & Abstraction can erase detail; exhaustive retention limits compression and raises cost. \\
\FMBetLabel{Empirical test} & Improved transfer to unseen entities, schemas, tasks, and consumers. \\
\end{tabularx}
\end{FMBet}

\textbf{Example.} After observing a transaction, a sequence model typically encodes it only to predict the next event. A state-oriented model may instead retain a representation of the transaction, update a transaction-trail summary, and update the states of the card, customer, merchant, their relationships, and relevant aggregates. These states remain available beyond the current context window and need not all be exposed to every consumer.

\textbf{Candidate mechanisms.} Observation encoding, state update, and selective state exposure provide an architecture-independent functional decomposition. To construct state representations, typed event encoders can preserve type-specific semantics~\citep{rusakov2026foundation}, while point-process components can preserve irregular timing~\citep{berghaus2026context}. Memory-augmented entity-centric attention can route observations to entity and relationship states~\citep{cheng2025graph}. To integrate new observations with prior state, event-driven memory mechanisms from temporal graphs~\citep{rossi2020temporal} could be adapted with type-aware learned update operators~\citep{li2026evolving}. To support multiple resolutions, multi-scale encoders can connect state across short and long temporal horizons~\citep{zhang2024multi}. Hierarchical aggregation with cross-level reconciliation constraints~\citep{wickramasuriya2024probabilistic} can relate state across entity, customer, business, and portfolio levels while enforcing relevant relational and accounting identities. Nested embeddings offer coarse-to-fine views of the same state representation~\citep{kusupati2022matryoshka}. These state representations need not be wholly latent: they may combine observed attributes, derived measures, learned features, and references to supporting evidence. Objectives that predict future state representations or transitions can encourage the current state to retain information needed for future dynamics, while cross-view alignment can promote agreement across sources and resolutions~\citep{zou2024interfacing,lu2025representation}. FMs can expose selective typed views designed to preserve transfer utility across tasks and latent spaces~\citep{moschella2022relative}; semantic-stability and disclosure constraints govern what those views may contain. 

\textbf{Key trade-off.} Explicit state can improve continuity and transfer but requires choices about what to persist, update, address, and expose. Greater abstraction may suppress numerical or source-specific detail, while retaining every event reproduces fragmentation and raises cost.

\textbf{Testable implications.} Explicit state should improve transfer to unseen entities, schemas, tasks, and consumers over implicit sequence representations, task-specific temporal graph models, and static entity embeddings at matched resource and adaptation budgets, not merely improve reconstruction or next-event prediction. Effective transfer from representations that persist no state beyond their context would weaken the bet.

\subsubsection{Learned representation interoperability}

\textbf{Research bet.} Given specialist financial FMs exposing semantically meaningful, multi-resolution state, a representation fabric should enable reuse by heterogeneous, evolving consumers through learned interoperability. Queryable contracts can declare a representation's intended meaning and constrain its use; learned components must determine which permitted state is useful, how to translate and compose it, and when to reject transfer. The bet is that these capabilities can support efficient consumer adaptation and independent model evolution without shared weights, a universal latent space, or separately engineered integration for every producer--consumer pair.

\begin{FMBet}
\small
\renewcommand{\arraystretch}{1.12}
\begin{tabularx}{\linewidth}{@{}p{0.16\linewidth}X@{}}
\FMBetLabel{Core claim} & Learned interoperability composes useful state and rejects unsupported transfer. \\
\FMBetLabel{Approaches} & State discovery; meta-learned adapters; compatibility assessment; routing and fusion. \\
\FMBetLabel{Central tension} & Independent producer optimization versus reliable consumer reuse. \\
\FMBetLabel{Empirical test} & Unseen pairings and upgrades need less adaptation than bespoke links. \\
\end{tabularx}
\end{FMBet}

\textbf{Example.} To reassess a corporate borrower, a credit-risk FM may request permitted transaction-flow state and liquidity state for the borrower’s issued securities. The fabric could discover entity- and time-aligned views from transaction and market FMs, translate and compose them with the credit FM’s native borrower state, and reject mappings whose entity correspondence, temporal aggregation, or semantics are insufficiently supported.

\textbf{Candidate mechanisms.} Discovery can identify applicable state views and feasible interoperability paths using representation contracts, semantic descriptors, and prior transfer evidence; it may occur offline when establishing reusable paths or at query time for a particular consumer and task. Directional translators can map discovered state into a consumer's representation space, trained from shared entities, aligned periods, cross-view objectives, or downstream supervision~\citep{moschella2022relative}. Shared translator families or meta-learned adapters, trained across existing producer--consumer pairs, may generalize these mappings to new pairs and versions from a small entity-aligned sample rather than a dedicated translator~\citep{chen2026transferring,jha2026harnessing}. Methods for backward-compatible alignment and studies of semantic convergence provide starting points for compatibility models that estimate semantic mismatch, information loss, and out-of-support transfer~\citep{ricci2024backward,son2025semantic}. Adaptive routing can use these estimates to choose among state views, resolutions, versions, translators, or composition paths according to expected task utility, confidence, cost, and latency~\citep{li2026routing}; gates can accept, downweight, or reject the resulting contributions. Consumer-side import adapters or fusion modules can compose accepted producer state with native state or with translated state from other specialists~\citep{pfeiffer2023modular}. Depending on available supervision and adaptation budgets, consumers may use imported state through in-context learning~\citep{qu2026tabiclv2}, parameter-efficient adaptation~\citep{han2024parameter}, or task-specific post-training~\citep{liu2026pre}, while producer cores remain fixed.

\textbf{Key trade-off.} Producer autonomy fosters domain-specific optimization but shifts much of the interoperability burden to the fabric. Strong alignment may improve composition while erasing specialist information, increasing cost, or exposing restricted attributes. Translator training also needs entity-aligned data, and permitted overlap may be limited. Fabric-side selection and aggregation can form task-relevant or coarser views when semantics and permissions allow, but cannot reliably reconstruct finer state from lossy summaries. Generalization across unseen models and versions may sacrifice fidelity relative to a bespoke pairwise adapter, while conservative compatibility gates may reject transfers that would have been useful.

\textbf{Testable implications.} Learned interoperability should enable useful transfer to previously unpaired consumers and across model upgrades with less adaptation and integration effort than bespoke pairwise integration. Compatibility estimates should identify unsupported or harmful transfer; failure on either condition would weaken the bet.

\subsection{Dynamics over financial state}

\subsubsection{Regime-conditioned state transitions}

\textbf{Research bet.} Separating durable state semantics from regime-dependent transition dynamics should improve prediction across recurring regimes. A stable state interface should represent the current financial condition, while an inferred regime belief determines how that state is likely to evolve. When learned dynamics recur, models should reactivate them without redefining state or retraining the full model; when dynamics are unsupported, adaptation should remain localized where possible.

\begin{FMBet}
\small
\renewcommand{\arraystretch}{1.12}
\begin{tabularx}{\linewidth}{@{}p{0.16\linewidth}X@{}}
\FMBetLabel{Core claim} & Separating state semantics from dynamics enables regime reuse and localized adaptation. \\
\FMBetLabel{Approaches} & Regime inference; transition experts; persistence priors; support and shift detection. \\
\FMBetLabel{Central tension} & Learned regimes may reflect state differences rather than distinct dynamics. \\
\FMBetLabel{Empirical test} & Recurring dynamics are reused without retraining or redefining state. \\
\end{tabularx}
\end{FMBet}

\textbf{Example.} In an order book, instrument identity and trading rules provide durable descriptors, while current spread, depth, and order imbalance form part of the evolving state. Under stress, the same state may evolve differently as cancellation intensity rises, depth replenishment slows, and price impact increases. A regime-aware FM would infer greater support for stressed transition dynamics and apply them to the same order-book state representation. It could forecast wider spreads, lower liquidity, and greater price impact, then return to ordinary dynamics as conditions normalize.

\textbf{Candidate mechanisms.} Regimes may take the form of discrete latent components, soft mixtures, or continuous embeddings rather than predefined labels~\citep{becker2019switching}. A jointly trained regime-assignment component can estimate regime weights or embeddings from recent state transitions, prediction residuals, and change-point evidence, while transition experts, adapters, or regime embeddings parameterize the corresponding dynamics over a shared state representation. Persistence-aware priors can encourage temporally coherent assignments rather than rapid switching in response to transient noise~\citep{fox2011sticky}. Multi-horizon objectives and latent state-space or generative transition models can learn these representations and dynamics across timescales~\citep{xu2025deep,gu2023mamba,dey2026trust}. Sharing the state interface across regime-conditioned transitions and imposing cross-regime consistency can encourage state semantics to remain stable as transition behaviour changes. Support estimation and shift detection can distinguish recurring from unsupported dynamics~\citep{truong2020selective}; the latter may require a localized transition component, while failures shared across regimes may instead call for revising the state representation.

\textbf{Key trade-off.} Too many regime components can absorb noise, while too few can conflate distinct dynamics or miss structural change. Detection that is too sensitive produces false switching, whereas conservative detection delays adaptation. The factorization is also not automatically identifiable: state may absorb regime information, regime variables may encode current condition rather than dynamics, and transition components may specialize by entity instead of behaviour. Strong invariance constraints may incorrectly treat changing relationships as durable structure.

\textbf{Testable implications.} When previously learned regimes recur, a regime-conditioned model should match periodically retrained regime-blind models without requiring parameter updates, while preserving the meaning and downstream usability of its state across regimes. For genuinely unsupported dynamics, updating or adding a localized transition component should be more effective and cause less forgetting than retraining the full model. Failure on either comparison, or evidence that state must be redefined across regimes, would weaken the bet.

\subsubsection{Action-conditioned state transitions}

\textbf{Research bet.} The credible path toward financial world models starts with controlled, action-conditioned transition models, not end-to-end simulators trained only on observational histories. Such transitions should represent both an action's direct effects and relevant behavioural responses, since deploying a policy can change the process being modelled~\citep{lucas1976econometric}. Claims should advance through an evidence ladder: from observational forecasting \(p(s_{t+1}\mid s_t)\), to conditional simulation \(p(s_{t+1}\mid s_t,a_t)\), then to off-policy evaluation, and ultimately to causal intervention \(p(s_{t+1}\mid s_t,\operatorname{do}(a_t))\) when identification is defensible.

\begin{FMBet}
\small
\renewcommand{\arraystretch}{1.12}
\begin{tabularx}{\linewidth}{@{}p{0.16\linewidth}X@{}}
\FMBetLabel{Core claim} & Action-conditioned transitions enable simulation and a graduated path to causal claims. \\
\FMBetLabel{Approaches} & Action-conditioned models; rollout objectives; off-policy evaluation; response models. \\
\FMBetLabel{Central tension} & Plausible simulations can mask wrong dynamics; conditioning alone is not causal evidence. \\
\FMBetLabel{Empirical test} & Improved prediction within support; stronger claims require defensible identification. \\
\end{tabularx}
\end{FMBet}

\textbf{Example.} A conventional portfolio model may forecast returns or risk for current holdings, or perturb those holdings and recompute its prediction. An action-conditioned FM would instead condition explicitly on a proposed rebalance or hedge action. It would update the portfolio state—including holdings, exposures, and concentration—as implied by that action, then simulate possible returns, execution costs, and liquidity as markets respond. These are conditional scenarios, not automatically causal estimates.

\textbf{Candidate mechanisms.} History-based transition models and input-perturbation analyses provide forecasting baselines but do not distinguish controllable actions from observed features. Action-conditioned transition and world models can encode a proposed action separately from current state and generate conditional trajectories~\citep{hansen2024td,xiang2025pan}. Training can combine one-step conditional prediction with multi-step rollout objectives, constraints on action-implied state changes and accounting consistency, and support-aware regularization against confident extrapolation. Behaviour-policy estimation and support diagnostics can identify where logged data permit stronger evaluation; within that support, off-policy evaluation methods can estimate outcomes under alternative policies~\citep{yu2024two,marmarelis2025off}. Randomized or quasi-experimental variation, structural assumptions, and validated simulators can support causal intervention claims. Separately, performative-prediction frameworks characterize how deploying a policy can change the data-generating process~\citep{hardt2025performative}; explicit participant-response models, game-theoretic or multi-agent simulators, and adversarial or best-response training can represent such adaptive behaviour. Uncertainty about support, transition dynamics, and participant response should remain explicit.

\textbf{Key trade-off.} More expressive simulators can generate plausible scenarios while concealing misspecified dynamics. Action conditioning alone does not identify causal effects, and modelling behavioural responses introduces assumptions about participants' objectives and information. Stronger claims narrow the conditions for which adequate support exists.

\textbf{Testable implications.} Action conditioning should improve prediction for held-out action--state combinations within the available support. Models of participant response should improve on action-conditioned models that assume historical responses remain fixed when repeated deployment changes how participants react. Off-policy or causal claims require corresponding identification assumptions and evidence; plausible trajectories alone do not suffice.

\subsection{Cross-cutting research programs}

\subsubsection{Learning from constrained financial data}

\textbf{Research bet.} Financial FMs should jointly optimize which evidence they consume and how evidence from different sources is combined under constraints on access, volume, context, and historical coverage. During pretraining and inference, selection should prioritize permitted evidence, including informative non-events, expected to alter or disambiguate financial state rather than rely only on similarity, recency, or salience. Across training stages, curricula should combine observed, privacy-transformed, distributed, and synthetic evidence to balance empirical grounding, structural coverage, and permitted use.

\begin{FMBet}
\small
\renewcommand{\arraystretch}{1.12}
\begin{tabularx}{\linewidth}{@{}p{0.16\linewidth}X@{}}
\FMBetLabel{Core claim} & State-informed selection and curricula preserve capability under data constraints. \\
\FMBetLabel{Approaches} & Adaptive sampling; temporal aggregation; state-aware retrieval; synthetic--real curricula. \\
\FMBetLabel{Central tension} & Selection can miss subtle signals; synthetic data inherit generator assumptions. \\
\FMBetLabel{Empirical test} & Maintained task quality with less context; valid, privacy-preserving synthetic coverage. \\
\end{tabularx}
\end{FMBet}

\textbf{Example.} A lender reassessing credit risk after a repayment-policy change may have billions of repetitive account events, restricted income and transaction records, and few historical outcomes under the new policy. A data-constrained FM could prioritize permitted observations that reveal changes in repayment capacity, aggregate low-information balance records, and use synthetic histories to supplement policy conditions poorly covered by observed data. At inference, it would retrieve only the events and relationships expected to clarify the customer's current state.

\textbf{Candidate mechanisms.} During pretraining, adaptive sampling, temporal aggregation, state-change detection, meta-learning, and importance weighting can prioritize informative observations and compress repetitive evidence~\citep{gu2025data,yu2026group,calian2026datarater}. During inference, hierarchical memory, graph retrieval, and temporal routing can select permitted entities, relationships, events, and periods according to their expected effect on state or uncertainty~\citep{deng2026influence}. A separate design choice is the composition and ordering of training evidence. Candidate curricula~\citep{garg2025real} include synthetic structural pretraining, synthetic pretraining followed by grounding on restricted or privacy-transformed real evidence, joint synthetic--real training, learning from unpooled data, and targeted augmentation of sparsely observed transitions~\citep{elsheikh2026privacy,schlegel2025generating}. Provenance-aware, use-scoped checkpoints can keep observed, transformed, and generated evidence distinguishable while retaining source classes, transformations, generator ancestry, approvals, and adaptation history.

\textbf{Key trade-off.} Aggressive selection reduces cost and noise but can discard weak or distant signals; broad context preserves optionality at potentially prohibitive cost. Synthetic supplementation trades broader coverage against dependence on generator assumptions and weaker grounding in observed evidence.

\textbf{Testable implications.} State-informed selection should maintain downstream quality while using less data and context than generic retrieval or fixed-window baselines, and removing selected evidence should alter relevant state or uncertainty. Synthetic supplementation should improve coverage of sparsely observed transitions without sacrificing privacy or structural validity.

\subsubsection{Uncertainty and tail-aware state}

\textbf{Research bet.} Financial FMs should learn state representations that retain information about infrequent states, extreme observed quantities, and unusual transitions that may matter during specialization, while explicitly representing uncertainty and evidential support. Uncertainty about inferred state, regime, translation, and future outcomes should remain distinguishable and propagate through composition.

\begin{FMBet}
\small
\renewcommand{\arraystretch}{1.12}
\begin{tabularx}{\linewidth}{@{}p{0.16\linewidth}X@{}}
\FMBetLabel{Core claim} & Separating uncertainties and modelling extremes supports reliable downstream use. \\
\FMBetLabel{Approaches} & Distributional state; shift and support detection; tail objectives; uncertainty propagation. \\
\FMBetLabel{Central tension} & Tail-focused learning can distort calibration; conservative propagation can reduce utility. \\
\FMBetLabel{Empirical test} & Improved tail estimates without ordinary-state loss; calibration through composition. \\
\end{tabularx}
\end{FMBet}

\textbf{Example.} A conventional risk model may return one default score even for a newly observed counterparty. An uncertainty- and tail-aware FM would separately represent how weakly the current counterparty state is supported, uncertainty about the prevailing regime, translation loss in imported exposure state, the likelihood of default and the severity of resulting loss or contagion trajectories. A consumer could then widen its loss distribution, downweight imported exposure state, or decline to issue a confident estimate when evidence or translation support is inadequate.

\textbf{Candidate mechanisms.} To represent uncertainty where it arises, producers can maintain uncertainty-aware latent states and attach distributional representations, ensembles, or prediction sets to exported state~\citep{zhang2026uncertainty,zhou2025conformal}. Shift and support detectors can identify states, regimes, or inputs that are weakly covered by training evidence~\citep{liu2026rag}. To improve tail behaviour, distributionally robust objectives can focus learning on specified adverse regions~\citep{shi2024distributionally}; extreme-value components can model tail severity, while quantile and expected-shortfall objectives can target the distribution of extreme losses~\citep{ns2023extreme,zhang2026resga}. When state crosses model boundaries, translators should preserve producer uncertainty while separately estimating semantic mismatch and translation loss~\citep{ling2024bayesian}. Composition should retain the source and type of uncertainty rather than collapse heterogeneous qualifications into a single confidence score. Consumers can marginalize over, downweight, reject, or escalate uncertain imported state and carry the resulting uncertainty into task outputs. Calibration-aware post-training can balance downstream decision quality with calibrated confidence during specialization~\citep{yaldiz2026balancing}.

\textbf{Key trade-off.} Greater tail sensitivity can improve detection of consequential extremes but increase false alarms or distort ordinary-state calibration. Conservative uncertainty propagation may make composed outputs uninformative, while aggressive aggregation may hide unsupported specialist or translated state.

\textbf{Testable implications.} Uncertainty-aware state and tail-sensitive specialization should improve calibration, unsupported-state detection, rare-event detection, severity ranking, and stressed-trajectory coherence relative to average-likelihood training alone, without unacceptable degradation under ordinary conditions. Uncertainty should remain calibrated through composition: it should increase under missing evidence, novel regimes, and lossy or conflicting transfer, but may decrease when independently informative state is successfully combined.

\subsubsection{Evidence-grounded explanation of state composition}

\textbf{Research bet.} Explanations grounded in interpretable state, provenance-aware evidence, and computational state traces will be more faithful and auditable than post-hoc narratives generated from an output alone. A trace should connect selected evidence, inferred state updates, translations, routing decisions, assumptions, uncertainty propagation, and specialist contributions that materially produced a composed output.

\begin{FMBet}
\small
\renewcommand{\arraystretch}{1.12}
\begin{tabularx}{\linewidth}{@{}p{0.16\linewidth}X@{}}
\FMBetLabel{Core claim} & Material state-to-output traces make explanations faithful and auditable. \\
\FMBetLabel{Approaches} & Concept-aligned state; provenance; intervention tests; entitlement-safe explanations. \\
\FMBetLabel{Central tension} & Richer traces improve auditability but raise disclosure, compute, and cognitive costs. \\
\FMBetLabel{Empirical test} & Material sensitivity, immaterial stability, and preserved validity and entitlements. \\
\end{tabularx}
\end{FMBet}

\textbf{Example.} A post-hoc explanation may identify influential features or semantic groups behind a credit-limit recommendation without tracing the composed computation. A trace-grounded explanation would connect permitted payment and utilization events to inferred customer-state changes, translated macroeconomic state, information lost in translation, and recommendation uncertainty, while verifying materiality by showing how the proposed limit changes when late-payment updates or the imported state are removed.

\textbf{Candidate mechanisms.} Concept bottlenecks can expose concept-aligned components of specialist state~\citep{koh2020concept}. Pairing such state with task heads whose dependencies are transparent can make parts of the state-to-output pathway interpretable by design~\citep{chattopadhyay2022interpretable,yordanov2026prototype}. Influence attribution, component ablation, and counterfactual intervention can test which state changes and computational pathways materially affected an output~\citep{hill2026hierarchical,zhu2026faithfulness}. Provenance-aware traces~\citep{wang2026agent} can connect these pathways to timestamped evidence and uncertainty, establishing lineage without by themselves demonstrating materiality. For composed outputs, such traces should additionally identify state and translator versions, translation and routing decisions, and imported specialist contributions. An explanation layer can then selectively expose an entitlement-safe subset of supported observations, inferences, contributions, and limitations, associate its claims with trace or intervention evidence, and communicate relevant uncertainty~\citep{raman2023bayesian}.

\textbf{Key trade-off.} Interpretable-by-design constraints may limit representation capacity. Richer traces can improve validation and contestability but may expose restricted information, increase computation, or overwhelm users with immaterial detail; selective explanations risk omitting weak but consequential contributions.

\textbf{Testable implications.} Trace-grounded explanations should change consistently under material interventions, remain stable under immaterial perturbations, and preserve provenance, temporal validity, and entitlements through composition. Failure to track changes in state, translation, or specialist contribution would weaken the claim of greater faithfulness.

\subsection{Scope and alternatives}

\paragraph{Research horizon.} Near-term work should establish whether persistent state and learned interoperability deliver reusable capability under leakage-resistant evaluation, with uncertainty and traceability tested throughout. Efficient downstream specialization and continual adaptation are enabling capabilities evaluated within these bets rather than separate research directions. Regime- and action-conditioned simulation can build on this foundation; general financial world models, strategic multi-agent simulation, and autonomous decision making remain longer-horizon ambitions.

\paragraph{Agentic orchestration.} Agentic orchestration is a downstream layer rather than part of the fabric: the fabric makes heterogeneous state discoverable, compatible, and governable, whereas an orchestrator plans sequential, multi-step workflows and selects the next specialist, tool, or action based on intermediate results. This extension raises distinct questions about bounded authority, permission-aware routing, failure recovery, and traces connecting orchestration decisions to specialist outputs and evidence~\citep{wang2026agent}.

\paragraph{Feedback-driven improvement.}
As specialists update from new observations over time, they must retain valid prior knowledge~\citep{guo2025comprehensive} while accounting for selective and delayed feedback shaped by which outcomes their decisions make observable. More autonomous forms of self-improvement, including models that generate their own feedback~\citep{qu2024recursive}, are a distinct, longer-horizon program beyond the scope of the representation fabric.

\paragraph{Inference constraints.} Latency, throughput, and state freshness matter especially in real-time fraud detection and trading. Tight response budgets may favour context selection, asynchronously updated state, precomputed interoperability paths, and lightweight task heads, with expensive composition and simulation off the critical path. Serving architectures are beyond this agenda.

\begin{FMDefinition}
\paragraph{Design choices.} Domain-specialized FMs that keep state implicit and treat actions as observed inputs offer a simpler alternative,  but do not expose reusable state for cross-model transfer or explicit transition models for intervention analysis. For interoperability and reuse across financial models, alternative designs make different trade-offs. A monolithic FM avoids translation but couples data, updates, and deployment; disconnected specialists preserve independence but reuse only outputs; pairwise adapters provide tailored transfer but multiply across pairs and versions; and agentic orchestration may suffice when output-level coordination is enough. The representation fabric instead targets state-level reuse across independently evolving specialists, accepting additional interoperability complexity to discover, translate, compose, and assess heterogeneous state. Its value should be assessed along the performance--reuse--governance frontier. 
\end{FMDefinition}

% =============================================================================
\section{How Should Progress Be Evaluated?}
\label{sec:evaluation}
% =============================================================================

Evaluation should test both generalization beyond entities and periods seen during development and reuse across new tasks and model combinations with limited adaptation. Random train--test splits and average scores can obscure these capabilities by mixing future and past information, repeating entities across splits, and underweighting consequential regimes and outcomes ~\citep{de2018advances, tan2021critical}. Tests should therefore mirror each claimed capability and report task quality separately from reuse value: progress may mean meeting stringent quality requirements or meeting practical requirements with substantially lower data, adaptation, integration, or governance cost. ~\autoref{tab:evaluation-framework} summarizes the evaluation principles; constructing credible tasks, datasets, and metrics remains part of the agenda.

\paragraph{Task performance and transfer must be evaluated together.}
Task-specific models, modality-specific FMs, and structured financial FMs should each be evaluated using fixed protocols, information, and adaptation budgets, with prespecified quality, calibration, and consequential-error criteria. Unseen entities, tasks, products, and regimes test distinct forms of transfer and should not collapse into one score. Results should identify what transfers and report a performance--reuse frontier: neither task performance without reuse nor low-cost sharing with unacceptable degradation validates the fabric.

\paragraph{Representation interoperability must generalize beyond fixed model pairs.}
Evaluation should hold out consumers, tasks, and model versions and assess learned interoperability alongside native-state consumption, pairwise adapters, universal latent spaces, score exchange, and disconnected specialists. Compatibility estimates should predict degradation under semantic mismatch, missing resolution, distribution shift, and producer replacement, enabling rejection or downweighting. Report these AI tests separately from permissions, lineage, and version enforcement.

\paragraph{Chronology and regime change are first-class splits.}
Evaluations must respect when information became available, including revisions and delayed reports. Training, adaptation, validation, and testing should be chronological, with entity overlap and relationship paths documented to expose leakage. Cross-regime tests ~\citep{arnott2018backtesting, cabral2026non} should distinguish drift from structural change, report both adaptation and retention, and remain robust across defensible regime definitions. For unseen dynamics, localized expansion should be compared with full-model retraining.

\iffalse
\begin{FMOverview}{Evaluation at a glance}
\small
\begin{tabularx}{\linewidth}{@{}>{\bfseries}p{0.26\linewidth}X@{}}
Task performance & Meets prespecified quality, calibration, and consequential-error criteria? \\
Adaptation efficiency & New tasks or domains with fewer labels, updates, and deployment effort? \\
Cross-model composition & New capabilities from combining independently trained specialists? \\
Independent evolution & Add, replace, or version specialists without system-wide retraining? \\
Trustworthiness & Chronological, calibrated, explainable, entitlement-preserving, and robust? \\
\end{tabularx}
\end{FMOverview}
\fi

\begin{table*}[t]
	\centering
	\small
	\captionsetup{justification=centering}
	\caption{Capability-oriented evaluation for structured financial foundation models.}
	\label{tab:evaluation-framework}
    \renewcommand{\arraystretch}{1.5}
	\begin{tabular}{p{0.24\textwidth}p{0.31\textwidth}p{0.35\textwidth}}
		\toprule
		\textbf{Capability} & \textbf{Test design} & \textbf{Evidence of progress} \\
		\midrule
		Task quality & Compare models under the same task-specific protocol & Meets task-specific quality and risk tolerances \\
		Transfer and adaptation & Hold out entities, tasks, schemas, or regimes & Required quality with less data or retraining \\
		Dynamics, actions \& tails & Test shifts, interventions, and stress chronologically & Retained regimes; warranted interventions; calibrated tails \\
		Information efficiency & Limit resources; compare evidence selection and data curricula & Quality under constraints; valid, privacy-preserving synthesis \\
		Interoperability & Hold out consumers and versions; inject incompatible state & Useful cross-model state use; reliable compatibility scores \\
		Trustworthiness & Perturb evidence and components; verify traces and access controls & Faithful and robust explanations; preserved validity and entitlements \\
		%System-level value & Compare with monolithic and agent-coordinated alternatives & Better performance--reuse--cost frontier \\
		\bottomrule
	\end{tabular}
\end{table*}

\paragraph{Intervention claims require stronger evidence.}
Evaluation suites should state whether evidence supports forecasting, conditional simulation, off-policy generalization, strategic response, or causal intervention ~\citep{pearl2016causal, gische2021forecasting}. Strategic-response tests should vary deployed policies and include adaptive or adversarial participant behaviour; counterfactual claims require experimental variation, validated simulators, known processes, or stated identification assumptions. Synthetic environments and simulators can provide controlled test beds with known interventions and regimes outside training, enabling tests of generalization.

\paragraph{Tails require conditional and joint reporting.}
Report ordinary, rare, stressed, and shifted conditions jointly ~\citep{aikman2024stress}, including calibration, unsupported inputs, severity ranking, scenario coherence, and ordinary--tail trade-offs. Tests should also verify that uncertainty responds appropriately to missing evidence, novel regimes, translation loss, and composition. Stress transformations and synthetic environments may supplement historical backtests only with explicit assumptions and preserved financial relationships and accounting identities.

\paragraph{Explanations must be tested for faithfulness and robustness.}
Perturbation, intervention, and trace-based tests should verify appropriate sensitivity and invariance as evidence, state updates, assumptions, uncertainty, and specialist contributions are varied~\citep{chan2022comparative, agarwal2022rethinking}. Explanations should distinguish observed data from inferred state and generated scenarios, remain coherent when specialists are composed or replaced, and preserve provenance, temporal validity, and entitlements.

\paragraph{Systems must be evaluated under realistic information constraints.}
End-to-end evaluation should constrain pretraining data, context, latency, compute, monetary cost, and access. Real-only, synthetic-first, transformed-real, and mixed training should be compared under matched budgets and on held-out real data, with synthetic supplementation assessed for privacy and structural validity. Performance--budget curves and ablations should measure efficiency and the contribution of selected evidence. Every suite should specify the task, transfer unit, chronological boundary, reference systems, metrics, and failure criterion.

\begin{FMDefinition}
\paragraph{The complete paradigm must improve on alternative system designs.}
End-to-end evaluation should compare the proposed paradigm with monolithic and disconnected specialists, pairwise adapters, and agent-coordinated specialists under matched task, information, adaptation, integration, and governance budgets, including existing retrieval, simulation, and downstream controls. Explicit state, localized dynamics, and learned interoperability justify their added complexity only if they deliver greater reuse under practical performance and governance constraints.
\end{FMDefinition}

% =============================================================================
\section{Conclusion}
\label{sec:conclusion}
% =============================================================================

Structured financial foundation models should not be defined by scale, a single architecture, or the breadth of pretraining. Their value lies in whether they recover reusable state from heterogeneous, asynchronous, and permissioned data, and make it useful across tasks, regimes, and independently evolving specialists. We argue for domain-specialized FMs that expose meaningful financial state, connected by a learned representation fabric that discovers, translates, and assesses state across model boundaries without bespoke integration for every pair. Operational controls govern these exchanges; generalizable representation interoperability is the central AI claim.

This proposal is a testable research agenda, not a deployment blueprint. Progress requires accepted task quality and efficient transfer under chronological, cross-regime, tail, and action-conditioned evaluation; composition that preserves semantics, uncertainty, provenance, and entitlements; and explanations faithful to the evidence and state changes behind an output. The broader structured-financial-FM paradigm is validated when reusable capability survives these constraints; scale and domain-specific pretraining provide important foundations but do not by themselves establish the full paradigm.

% =============================================================================
\section*{Acknowledgments}
% =============================================================================
This paper was prepared for informational purposes by the Artificial Intelligence Research group of JPMorgan Chase \& Co. and its affiliates ("JP Morgan") and is not a product of the Research Department of JP Morgan. JP Morgan makes no representation and warranty whatsoever and disclaims all liability, for the completeness, accuracy or reliability of the information contained herein. This document is not intended as investment research or investment advice, or a recommendation, offer or solicitation for the purchase or sale of any security, financial instrument, financial product or service, or to be used in any way for evaluating the merits of participating in any transaction, and shall not constitute a solicitation under any jurisdiction or to any person, if such solicitation under such jurisdiction or to such person would be unlawful. © 2026 JP Morgan Chase \& Co. All rights reserved.

\bibliography{references}
\bibliographystyle{IEEEtranN}

%\appendix
%\input {sec_appendix}

\end{document}